\documentclass[11pt,letterpaper]{article}
\usepackage{naaclhlt2016}
\usepackage{times}
\usepackage{latexsym}

\usepackage{amsmath}
\usepackage{amsfonts}
\usepackage{float}
\usepackage{multirow}
\usepackage[utf8]{inputenc}
\usepackage[T1]{fontenc}
\usepackage[spanish, german, french, italian, dutch, polish, romanian, portuguese, russian, english]{babel}

\makeatletter
\newcommand{\@BIBLABEL}{\@emptybiblabel}
\newcommand{\@emptybiblabel}[1]{}
\makeatother
\usepackage[hidelinks]{hyperref}
\usepackage{url}

\usepackage[final]{graphicx}
\usepackage[ruled]{algorithm2e}

\def\BState{\State\hskip-\ALG@thistlm}
\usepackage{url}

\naaclfinalcopy 
\def\naaclpaperid{339} 

\title{Bridge Correlational Neural Networks for Multilingual Multimodal Representation Learning}

\author{Janarthanan Rajendran \\
  IIT Madras, India. \\
  {\scriptsize{\tt rsdjjana@gmail.com}} \\
  \And
  Mitesh M. Khapra \\
  IBM Research India. \\
  {\scriptsize{\tt mikhapra@in.ibm.com}} \\
  \And
  Sarath Chandar\\
  University of Montreal.\\
  {\scriptsize{\tt apsarathchandar@gmail.com}} \\
  \And
  Balaraman Ravindran\\
  IIT Madras, India.\\
  {\scriptsize{\tt ravi@cse.iitm.ac.in}}\\
  }

\date{}

\begin{document}

\maketitle

\begin{abstract}
 
Recently there has been a lot of interest in learning common representations for multiple views of data. Typically, such common representations are learned using a parallel corpus between the two views (say, 1M images and their English captions). In this work, we address a real-world scenario where no direct parallel data is available between two views of interest (say, $V_1$ and $V_2$) but parallel data is available between each of these views and a pivot view ($V_3$). We propose a model for learning a common representation for  $V_1$, $V_2$ and $V_3$ using only the parallel data available between $V_1V_3$ and $V_2V_3$. The proposed model is generic and even works when there are $n$ views of interest and only one pivot view which acts as a bridge between them.
There are two specific downstream applications that we focus on (i) transfer learning between languages $L_1$,$L_2$,...,$L_n$ using a pivot language $L$ and (ii) cross modal access between images and a language $L_1$ using a pivot language $L_2$.
Our model achieves state-of-the-art performance in multilingual document classification on the publicly available multilingual TED corpus and promising results in multilingual multimodal retrieval on a new dataset created and released as a part of this work.
 
 \end{abstract}

\section{Introduction}
The proliferation of multilingual and multimodal content online has ensured that multiple views of the same data exist. For example, it is common to find the same article published in multiple languages online in multilingual news articles, multilingual wikipedia articles, \textit{etc}. Such multiple views can even belong to different modalities. For example, images and their textual descriptions are two views of the same entity. Similarly, audio, video and subtitles of a movie are multiple views of the same entity.

Learning common representations for such multiple views of data will help in several downstream applications. For example, 
learning a common representation for images and their textual descriptions could help in finding images which match a given textual description. Further, such common representations can also facilitate transfer learning between views. For example, a document classifier trained on one language (view) can be used to classify documents in another language by representing documents of both languages in a common subspace.

Existing approaches to common representation learning \cite{ngiam11,KlementievA2012,aps2,aps,dcca,wang} except \cite{HermannK2014} typically require parallel data between all views. However, in many real-world scenarios such parallel data may not be available. For example, while there are many publicly available datasets containing images and their corresponding English captions, it is very hard to find datasets containing images and their corresponding captions in Russian, Dutch, Hindi, Urdu, \textit{etc.} In this work, we are interested in addressing such scenarios. More specifically, we consider scenarios where we have $n$ different views but parallel data is only available between each of these views, and a pivot view. In particular, there is no parallel data available between the non-pivot views. 

To this end, we propose Bridge Correlational Neural Networks (Bridge CorrNets) which learn aligned representations across multiple views using a pivot view. We build on the work of \cite{corrnet} but unlike their model, which only addresses scenarios where direct parallel data is available between two views, our model can work for $n$($\geq$2) views even when no parallel data is available between all of them. Our model only requires parallel data between each of these $n$ views and a pivot view. During training, our model maximizes the correlation between the representations of the pivot view and each of the $n$ views. Intuitively, the pivot view ensures that similar entities across different views get mapped close to each other since the model would learn to map each of them close to the corresponding entity in the pivot view.

We evaluate our approach using two downstream applications. First, we employ our model to facilitate transfer learning between multiple languages using English as the pivot language. For this, we do an extensive evaluation using 110 source-target language pairs and clearly show that we outperform the current state-of-the art approach \cite{HermannK2014}. Second, we employ our model to enable cross modal access between images and French/German captions using English as the pivot view. For this, we created a test dataset consisting of images and their captions in French and German in addition to the English captions which were publicly available. To the best of our knowledge, this task of retrieving images given French/German captions (and vice versa) without direct parallel training data between them has not been addressed in the past. Even on this task we report promising results. Code and data used in this paper can be downloaded from \url{http://sarathchandar.in/bridge-corrnet}.

\section{Related Work}

Canonical Correlation Analysis (CCA) and its variants \cite{cca,rcca2,rcca3,rcca1,kcca} are the most commonly used methods for learning a common representation for two views. However, most of these models generally work with two views only. Even though there are multi-view generalizations of CCA \cite{mcca,tcca}, their computational complexity makes them unsuitable for larger data sizes. 

Another class of algorithms for multiview learning is based on Neural Networks. One of the earliest neural network based model for learning common representations was proposed in \cite{nlcca}. Recently, there has been a renewed interest in this field and several neural network based models have been proposed. For example, Multimodal Autoencoder  \cite{ngiam11}, Deep Canonically Correlated Autoencoder \cite{wang}, Deep CCA \cite{dcca} and Correlational Neural Networks (CorrNet) \cite{corrnet}. CorrNet performs better than most of the above mentioned methods and we build on their work as discussed in the next section.

One of the tasks that we address in this work is multilingual representation learning where the aim is to learn aligned representations for words across languages. Some notable neural network based approaches here include  the works of \cite{KlementievA2012,ZhouW2013,MikolovT2013,HermannK2014,Hermann2014ICLR,aps,soyer2015,gouws}. However, except for \cite{Hermann2014ICLR,HermannK2014}, none of these other works handle the case when parallel data is not available between all languages. Our model addresses this issue and outperforms the model of  \newcite{HermannK2014}.

The task of cross modal access between images and text addressed in this work comes under MultiModal Representation Learning where each view belongs to a different modality.  \newcite{ngiam11}  proposed an autoencoder based solution to learning common representation for audio and video. \newcite{JMLRv15srivastava14b} extended this idea to RBMs and learned common representations for image and text. Other solutions for image/text representation learning include \cite{DBLPjournals/corr/ZhengZL14,DBLPconf/cvpr/ZhengZL14,DBLPjournals/tacl/SocherKLMN14}. All these approaches require parallel data between the two views and do not address multimodal, multilingual learning in situations where parallel data is available only between different views and a pivot view.

In the past, pivot/bridge languages have been used to facilitate MT (for example, \cite{Wu07,Cohn07,Utiyama07,Nakov09}), transitive CLIR \cite{Ballesteros00,Lehtokangas08}, transliteration and transliteration mining \cite{Khapra10,Kumaran10,Khapra10a,Zhang11}. None of these works use neural networks but it is important to mention them here because they use the concept of a pivot language (view) which is central to our work.

\section{Bridge Correlational Neural Network}
In this section, we describe Bridge CorrNet which is an extension of the CorrNet model proposed by \cite{corrnet}. They address the problem of learning common representations between two views when parallel data is available between them. We propose an extension to their model which simultaneously learns a common representation for $M$ views when parallel data is available only between one pivot view and the remaining $M-1$ views. 

Let these views be denoted by $V_1,V_2,...,V_M$ and let $d_1,d_2,...,d_M$ be their respective dimensionalities. Let the training data be $\mathcal{Z} = \{z^i\}_{i=1}^{N}$ where each training instance contains only two views, \textit{i.e.}, $z^i = (v_j^i, v_M^i)$ where $j \in \{1,2,..,M{-1}\}$ and $M$ is a pivot view. To be more clear, the training data contains $N_1$ instances for which $(v_1^i, v_M^i)$ are available, $N_2$ instances for which $(v_2^i, v_M^i)$ are available and so on till $N_{M-1}$ instances for which $(v_{M-1}^i, v_M^i)$ are available (such that $N_1 + N_2 + ... + N_{M-1} = N$). We denote each of these disjoint pairwise training sets by $\mathcal{Z}_{_{1}}$, $\mathcal{Z}_{_{2}}$ to $\mathcal{Z}_{_{M-1}}$ such that $\mathcal{Z}$ is the union of all these sets. 

As an illustration consider the case when English, French and German texts are the three views of interest with English as the pivot view. As training data, we have $N_1$ instances containing English and their corresponding French texts and $N_2$ instances containing English and their corresponding German texts. We are then interested in learning a common representation for English, French and German even though we do not have any training instance containing French and their corresponding German texts. 

 \begin{figure}[h]
 \begin{center}
 \includegraphics[height=45mm,width=65mm]{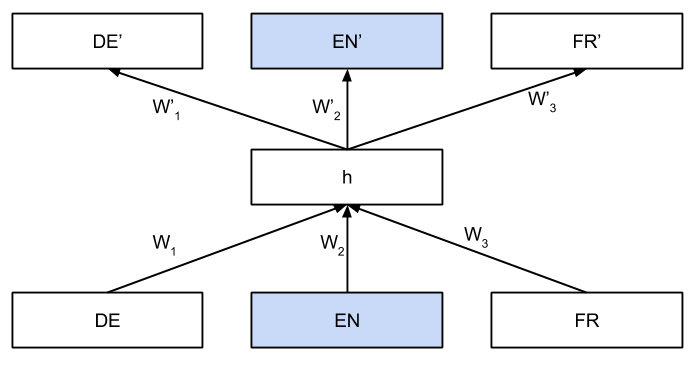}
 \caption{Bridge Correlational Neural Network. The views are English, French and German with English being the pivot view.}
 \label{f1}
 \end{center}
 \vskip -0.2in
 \end{figure} 

Bridge CorrNet uses an encoder-decoder architecture with a correlation based regularizer to achieve this. It contains one encoder-decoder pair for each of the $M$ views. For each view $V_j$, we have,
\begin{equation}
h_{V_j}(v_j) = f(W_j v_j + b)
\end{equation}
where $f$ is any non-linear function such as sigmoid or tanh, $W_j \in \mathbb{R}^{k \times d_j}$ is the encoder matrix for view $V_j$, $b \in \mathbb{R}^{k}$ is the common bias shared by all the encoders. We also compute a hidden representation for the concatenated training instance $z = (v_j, v_M)$ using the following encoder function:
\begin{equation}
h_{Z}(z) = f(W_j v_j + W_{_M} v_{_M} + b)
\end{equation}

\noindent In the remainder of this paper, whenever we drop the subscript for the encoder, then the encoder is determined by its argument. For example $h(v_j)$ means $h_{V_j}(v_j)$, $h(z)$ means $h_Z(z)$ and so on.

Our model also has a decoder corresponding to each view as follows:
\begin{equation}
g_{_{V_j}}(h) = p(W_j' h + c_j)
\end{equation}
where $p$ can be any activation function, $W_j' \in \mathbb{R}^{d_j \times k}$ is the decoder matrix for view $V_j$, $c_j \in \mathbb{R}^{d_j}$ is the decoder bias for view $V_j$. We also define $g(h)$ as simply the concatenation of $[g_{_{V_j}}(h), g_{_{V_M}}(h)]$. 

In effect, $h_{_{V_j}}(.)$ encodes the input $v_j$ into a hidden representation $h$ and then $g_{_{V_j}}(.)$ tries to decode/reconstruct $v_j$ from this hidden representation $h$. Note that $h$ can be computed using $h(v_j)$ or $h(v_M)$. The decoder can then be trained to decode/reconstruct both $v_j$ and $v_M$ given a hidden representation computed using any one of them. More formally, we train Bridge CorrNet by minimizing the following objective function:
\begin{small}
\begin{align}
\label{eq:objective}
\mathcal{J}_{\cal Z}(\theta) &= \sum_{i=1}^N  L(z^i,g(h(z^i)))  +  \sum_{i=1}^N L(z^i,g(h(v_{l(i)}^i))) \notag\\
&  + \sum_{i=1}^N L(z^i,g(h(v^i_{_M}))) - \lambda ~{\rm corr}(h(V_{l(i)}),h(V_M))  
\end{align}
\end{small}
where $l(i) = j$ if $z^i \in \mathcal{Z}_{_{j}}$ and the correlation term ${\rm corr}$ is defined as follows:
\begin{small}
\begin{equation}
 {\rm corr} = \frac{\sum_{i=1}^N (h(x^i) - \overline{h(X)})(h(y^i) - \overline{h(Y)})  }{\sqrt{\sum_{i=1}^N (h(x^i) - \overline{h(X)})^2 \sum_{i=1}^N (h(y^i) - \overline{h(Y)})^2    }}
\end{equation}
\end{small}
Note that $g(h(z^i))$ is the reconstruction of the input $z^i$ after passing through the encoder and decoder. $L$ is a loss function which captures the error in this reconstruction, $\lambda$ is the scaling parameter to scale the last term with respect to the remaining terms, $\overline{h(X)}$ is the mean vector for the hidden representations of the first view and $\overline{h(Y)}$ is the mean vector for the hidden representations of the second view. 

We now explain the intuition behind each term in the objective function. The first term captures the error in reconstructing the concatenated input $z^i$ from itself. The second term captures the  error in reconstructing both views given the non-pivot view, $v_{l(i)}^i$. The third term captures the  error in reconstructing both views given the pivot view, $v^i_{_M}$. Minimizing the second and third terms ensures that both the views can be predicted from any one view. Finally, the correlation term ensures that the network learns correlated common representations for all views. 

Our model can be viewed as a generalization of the two-view CorrNet model proposed in \cite{corrnet}. By learning joint representations for multiple views using disjoint training sets $\mathcal{Z}_{_{1}}$, $\mathcal{Z}_{_{2}}$ to $\mathcal{Z}_{_{M-1}}$ it eliminates the need for $^nC_2$ pair-wise parallel datasets between all views of interest.
The pivot view acts as a bridge and ensures that similar entities across different views get mapped close to each other since all of them would be close to the corresponding entity in the pivot view. 

Note that unlike the objective function of CorrNet \cite{corrnet}, the objective function of Equation \ref{eq:objective}, is a dynamic objective function which changes with each training instance. In other words, $l(i) \in \{1,2,..,M{-1}\}$ varies for each $i \in \{1,2,..,N\}$. For efficient implementation, we construct mini-batches where each mini-batch will come from only one of the sets $\mathcal{Z}_{_{1}}$ to $\mathcal{Z}_{_{M-1}}$. We randomly shuffle these mini-batches and use corresponding objective function for each mini-batch.


    

    

  

As a side note, we would like to mention that in addition to $\mathcal{Z}_{_{1}}$, $\mathcal{Z}_{_{2}}$ to $\mathcal{Z}_{_{M-1}}$ as defined earlier, if additional parallel data is available between some of the non-pivot views then the objective function can be suitably modified to use this parallel data to further improve the learning. However, this is not the focus of this work and we leave this as a possible future work.

\section{Datasets}
In this section, we describe the two datasets that we used for our experiments.

\subsection{Multlingual TED corpus}
\label{sec:ted_data}
\newcite{HermannK2014} provide a multilingual corpus based on the TED corpus for IWSLT
2013 \cite{cettoloEtAlEAMT2012}. It contains
English transcriptions of several talks from the TED conference and their translations in multiple languages. We use the parallel data between English and other languages for training Bridge Corrnet (English, thus, acts as the pivot langauge). \newcite{HermannK2014} also propose a multlingual document classification task using this corpus. The idea is to use the keywords associated with each talk (document) as class labels and then train a classifier to predict these classes. There are one or more such keywords associated with each talk but only the 15 most frequent keywords across all documents are considered as class labels. We used the same pre-processed splits\footnote{\url{http://www.clg.ox.ac.uk/tedcorpus}} as provided by \cite{HermannK2014}.
The training corpus consists of a total of 12,078 parallel documents distributed across 12 language pairs. 

\subsection{Multilingual Image Caption dataset}
\label{sec:mscoco_data}
The MSCOCO dataset\footnote{\url{http://mscoco.org/dataset/}} contains images and their English captions. On an average there are 5 captions per image. The standard train/valid/test splits for this dataset are also available online. However, the reference captions for the images in the test split are not provided. Since we need such reference captions for evaluations, we create a new train/valid/test of this dataset. Specifically, we take 80K images from the standard train split and 40K images from the standard valid split. We then randomly split the merged 120K images into train(118K), validation (1K) and test set (1K).

We then create a multilingual version of the test data by collecting French and German translations for all the 5 captions for each image in the test set. We use crowdsourcing to do this. We used the CrowdFlower
platform and solicited one French and one German translation for each of the 5000 captions using native speakers. We got each translation verified by 3 annotators. We restricted the geographical location of annotators based on the target language. We found that roughly 70\% of the French translations and 60\% of the German translations were marked as correct by a majority of the verifiers. On further inspection with the help of in-house annotators, we found that the errors were mainly syntactic and the content words are translated correctly in most of the cases. Since none of the approaches described in this work rely on syntax, we decided to use all the 5000 translations as test data.
This multilingual image caption test data (MIC test data) will be made publicly available\footnote{\url{http://sarathchandar.in/bridge-corrnet}} and will hopefully assist further research in this area.



\section{Experiment 1: Transfer learning using a pivot language}
From the TED corpus described earlier, we consider English transcriptions and their translations in 11 languages, \textit{viz.}, Arabic, German, Spanish, French, Italian, Dutch, Polish, Portuguese (Brazilian), Roman, Russian and Turkish.
Following the setup of \newcite{HermannK2014}, we consider the task of cross language learning between each of the $^{11}C_2$ non-English language pairs. The task is to classify documents in a language when no labeled training data is available in this language but training data is available in another language. This involves the following steps:

\noindent \textbf{1. Train classifier:} Consider one language as the source language and the remaining 10 languages as target languages. Train a document classifier using the labeled data of the source language, where each training document is represented using the hidden representation computed using a trained Bridge Corrnet model. As in \cite{HermannK2014} we used an averaged perceptron trained for 10 epochs as the classifier for all our experiments. The train split provided by \cite{HermannK2014} is used for training.

\noindent \textbf{2. Cross language classification:} For every target language, compute a hidden representation for every document in its test set using Bridge CorrNet. Now use the classifier trained in the previous step to classify this document. The test split provided by \cite{HermannK2014} is used for testing.



\subsection{Training and tuning Bridge Corrnet}
For the above process to work, we first need to train Bridge Corrnet so that it can then be used for computing a common hidden representation for documents in different languages. For training Bridge CorrNet, we treat English as the pivot language (view) and construct parallel training sets $\mathcal{Z}_{_{1}}$ to $\mathcal{Z}_{_{11}}$. Every instance in $\mathcal{Z}_{_{1}}$ contains the English and Arabic view of the same talk (document). Similarly, every instance in $\mathcal{Z}_{_{2}}$ contains the English and German view of the same talk (document) and so on. For every language, we first construct a vocabulary containing all words appearing more than 5 times in the corpus (all talks) of that language. We then use this vocabulary to construct a bag-of-words representation for each document. The size of the vocabulary ($|V|$) for different languages varied from 31213 to 60326 words. To be more clear, $v_1 = v_{{arabic}} \in \mathbb{R}^{|V|_{arabic}}$, $v_2 = v_{{german}} \in \mathbb{R}^{|V|_{german}}$ and so on. 


\begin{table*}[!t]
\centering
\begin{scriptsize}
\begin{tabular}{llllllllllll}
\hline
\multirow{2}{*}{\begin{tabular}[c]{@{}l@{}}Training\\ Language\end{tabular}} & \multicolumn{11}{c}{Test Language} \\ \cline{2-12}
 & Arabic & German & Spanish & French & Italian & Dutch & Polish & Pt-Br & Rom'n & Russian & Turkish \\ \hline
Arabic & \textbf{} & \textbf{0.662} & \textbf{0.654} & \textbf{0.645} & \textbf{0.663} & \textbf{0.654} & \textbf{0.626} & \textbf{0.628} & \textbf{0.630} & \textbf{0.607} & \textbf{0.644} \\
German & \textbf{0.920} & \textbf{} & \textbf{0.544} & \textbf{0.505} & \textbf{0.654} & \textbf{0.672} & \textbf{0.631} & \textbf{0.507} & \textbf{0.583} & \textbf{0.537} & \textbf{0.597} \\
Spanish & \textbf{0.666} & \textbf{0.465} & \textbf{} & \textbf{0.547} & \textbf{0.512} & \textbf{0.501} & \textbf{0.537} & \textbf{0.518} & \textbf{0.573} & \textbf{0.463} & \textbf{0.434} \\
French & \textbf{0.761} & \textbf{0.585} & \textbf{0.679} & \textbf{} & \textbf{0.681} & \textbf{0.646} & \textbf{0.671} & \textbf{0.650} & \textbf{0.675} & \textbf{0.613} & \textbf{0.578} \\
Italian & \textbf{0.701} & \textbf{0.421} & 0.456 & 0.457 &  & \textbf{0.530} & \textbf{0.442} & \textbf{0.491} & \textbf{0.390} & \textbf{0.402} & \textbf{0.499} \\
Dutch & \textbf{0.847} & \textbf{0.370} & \textbf{0.511} & \textbf{0.472} & \textbf{0.600} & \textbf{} & \textbf{0.536} & \textbf{0.489} & \textbf{0.458} & \textbf{0.470} & \textbf{0.516} \\
Polish & \textbf{0.533} & \textbf{0.387} & \textbf{0.556} & \textbf{0.535} & \textbf{0.536} & \textbf{0.454} & \textbf{} & \textbf{0.446} & \textbf{0.521} & \textbf{0.473} & \textbf{0.413} \\
Pt-Br & \textbf{0.609} & \textbf{0.502} & \textbf{0.572} & \textbf{0.553} & \textbf{0.548} & \textbf{0.535} & \textbf{0.545} & \textbf{} & \textbf{0.557} & \textbf{0.451} & \textbf{0.463} \\
Rom'n & \textbf{0.573} & \textbf{0.460} & \textbf{0.559} & \textbf{0.530} & \textbf{0.521} & \textbf{0.484} & \textbf{0.475} & \textbf{0.485} & \textbf{} & \textbf{0.486} & \textbf{0.458} \\
Russian & \textbf{0.755} & \textbf{0.460} & \textbf{0.537} & \textbf{0.437} & \textbf{0.567} & \textbf{0.499} & \textbf{0.550} & \textbf{0.478} & \textbf{0.475} & \textbf{} & \textbf{0.484} \\
Turkish & \textbf{0.950} & \textbf{0.373} & \textbf{0.480} & \textbf{0.452} & \textbf{0.542} & \textbf{0.544} & \textbf{0.585} & 0.297 & \textbf{0.512} & \textbf{0.412} & \textbf{} \\ \hline
\end{tabular}
\end{scriptsize}
\caption{F1-scores for TED corpus document classification results when training and testing on two
languages that do not share any parallel data. We train a Bridge CorrNet model on all en-L2 language pairs
together, and then use the resulting embeddings to train document classifiers in each language. These
classifiers are subsequently used to classify data from all other languages.}
 \label{table:ted_bridge}
\end{table*}

\begin{table*}[!t]
\centering
\begin{scriptsize}
\begin{tabular}{llllllllllll}
\hline
\multirow{2}{*}{\begin{tabular}[c]{@{}l@{}}Training\\ Language\end{tabular}} & \multicolumn{11}{c}{Test Language} \\\cline{2-12}
 & Arabic & German & Spanish & French & Italian & Dutch & Polish & Pt-Br & Rom'n & Russian & Turkish \\ \hline
Arabic &  & 0.378 & 0.436 & 0.432 & 0.444 & 0.438 & 0.389 & 0.425 & 0.42 & 0.446 & 0.397 \\
German & 0.368 &  & 0.474 & 0.46 & 0.464 & 0.44 & 0.375 & 0.417 & 0.447 & 0.458 & 0.443 \\
Spanish & 0.353 & 0.355 &  & 0.42 & 0.439 & 0.435 & 0.415 & 0.39 & 0.424 & 0.427 & 0.382 \\
French & 0.383 & 0.366 & 0.487 &  & 0.474 & 0.429 & 0.403 & 0.418 & 0.458 & 0.415 & 0.398 \\
Italian & 0.398 & 0.405 & \textbf{0.461} & \textbf{0.466} & \textbf{} & 0.393 & 0.339 & 0.347 & 0.376 & 0.382 & 0.352 \\
Dutch & 0.377 & 0.354 & 0.463 & 0.464 & 0.46 &  & 0.405 & 0.386 & 0.415 & 0.407 & 0.395 \\
Polish & 0.359 & 0.386 & 0.449 & 0.444 & 0.43 & 0.441 &  & 0.401 & 0.434 & 0.398 & 0.408 \\
Pt-Br & 0.391 & 0.392 & 0.476 & 0.447 & 0.486 & 0.458 & 0.403 &  & 0.457 & 0.431 & 0.431 \\
Rom'n & 0.416 & 0.32 & 0.473 & 0.476 & 0.46 & 0.434 & 0.416 & 0.433 &  & 0.444 & 0.402 \\
Russian & 0.372 & 0.352 & 0.492 & 0.427 & 0.438 & 0.452 & 0.43 & 0.419 & 0.441 &  & 0.447 \\
Turkish & 0.376 & 0.352 & 0.479 & 0.433 & 0.427 & 0.423 & 0.439 & \textbf{0.367} & 0.434 & 0.411 &  \\ \hline 
\end{tabular}
\end{scriptsize}
\caption{F1-scores for TED corpus document classification results when training and testing on two
languages that do not share any parallel data. Same procedure as Table \ref{table:ted_bridge}, but with DOC/ADD model in \protect\cite{HermannK2014}. 
}
 \label{table:blunsom}
\end{table*}

 \begin{table*}[!t]
 \centering
 \begin{scriptsize}
 \begin{tabular}{llllllllllll}
 \hline
 \multirow{2}{*}{Setting} & \multicolumn{11}{c}{Languages} \\\cline{2-12}
  & Arabic & German & Spanish & French & Italian & Dutch & Polish & Pt-Br & Rom'n & Russian & Turkish \\\hline
 Raw Data NB & \textbf{0.469} & \textbf{0.471} & 0.526 & \textbf{0.532} & 0.524 & 0.522 & 0.415 & 0.465 & \textbf{0.509} & 0.465 & \textbf{0.513} \\\hline
 DOC/ADD (Single) & 0.422 & 0.429 & 0.394 & 0.481 & 0.458 & 0.252 & 0.385 & 0.363 & 0.431 & \textbf{0.471} & 0.435 \\
 DOC/BI (Single) & 0.432 & 0.362 & 0.336 & 0.444 & 0.469 & 0.197 & 0.414 & 0.395 & 0.445 & 0.436 & 0.428 \\\hline
 DOC/ADD (Joint) & 0.371 & 0.386 & 0.472 & 0.451 & 0.398 & 0.439 & 0.304 & 0.394 & 0.453 & 0.402 & 0.441 \\
 DOC/BI (Joint) & 0.329 & 0.358 & 0.472 & 0.454 & 0.399 & 0.409 & 0.340 & 0.431 & 0.379 & 0.395 & 0.435 \\\hline
 Bridge CorrNet & 0.266 & 0.456 & \textbf{0.535} & 0.529 & \textbf{0.551} & \textbf{0.565} & \textbf{0.478} & \textbf{0.535} & 0.490 & 0.447 & 0.477 \\
 \hline
 \end{tabular}
 \end{scriptsize}
 \caption{: F1-scores on the TED corpus document classification task when training and evaluating on the
 same language. Results other than Bridge CorrNet are taken from \protect\cite{HermannK2014}.}
\label{table:mono}
\end{table*}

\begin{table*}
    \centering
    \includegraphics[width = \linewidth]{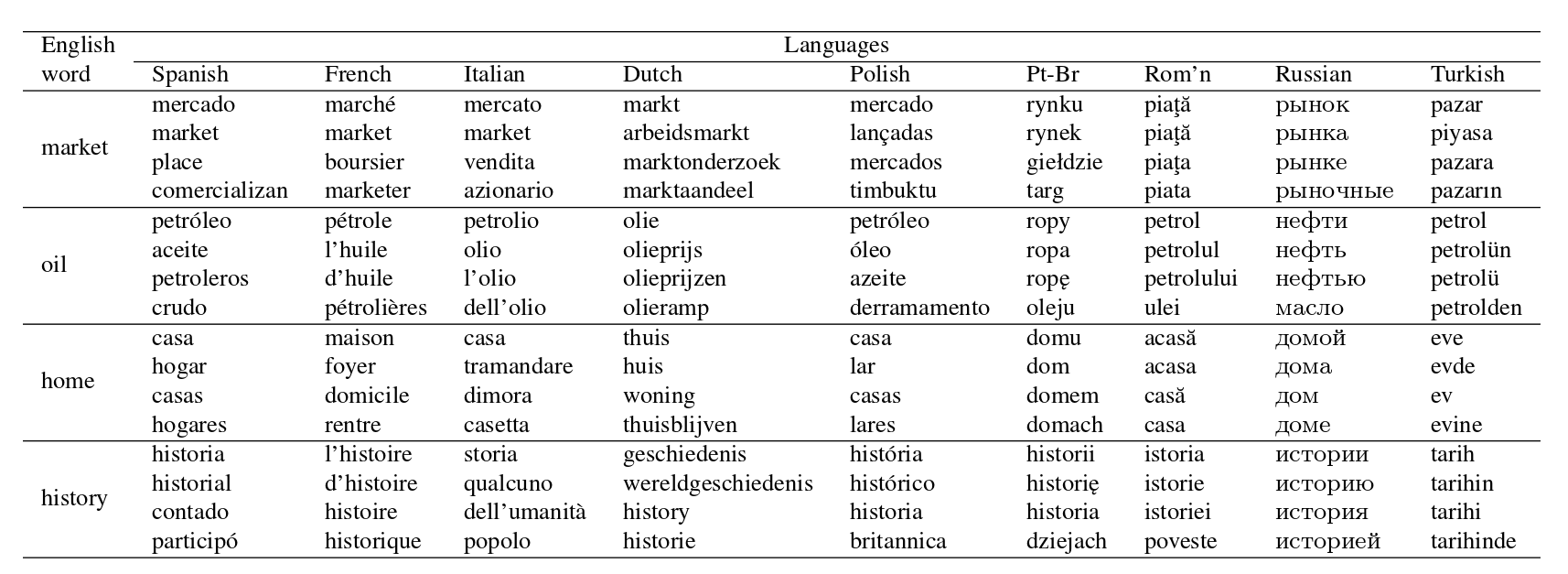}
    \caption{English words and their nearest neighbours in 9 languages (based on Euclidean distance).}
    \label{table:nearest_neighbors}
\end{table*}

We train our model for 10 epochs using the above training data $\mathcal{Z} = \{\mathcal{Z}_{_{1}}, \mathcal{Z}_{_{2}}, ..., \mathcal{Z}_{_{11}}\}$. We use hidden representations of size $D=128$, as in \cite{HermannK2014}. Further, we used stochastic gradient descent with mini-batches of size $20$. Each mini-batch contains data from only one of the  $\mathcal{Z}_{_{i}}s$. We get a stochastic estimate for the correlation term in the objective function using this mini-batch.
The hyperparameter $\lambda$ was tuned to each task using a training/validation split for the source language and using the performance on the validation set of an averaged perceptron trained on the training set (notice that this corresponds to a monolingual classification experiment, since the general assumption is that no labeled data is available in the target language). 

\subsection{Results}
Before presenting the results for our cross language classification experiment, we would first like to give a qualitative feel for the representations learned using Bridge CorrNet. For this, we randomly select a few English words and find their nearest neighbors in different languages based on the representations learned using Bridge CorrNet. These English words and their neighbors are shown in Table \ref{table:nearest_neighbors}. In almost all the cases the nearest neighbors of the English words turn out to be their exact translations or highly semantically related words. Also, we observed that the representations of translation pairs in non-English languages (say, French and German) are also transitively close to each other due to the pivot language.

We now present the results of our cross language classification task in Table \ref{table:ted_bridge}. Each row corresponds to a source language and each column corresponds to a target language. We report the average F1-scores over all the 15 classes. We compare our results with the best results reported in \cite{HermannK2014} (see Table \ref{table:blunsom}). Out of the 110 experiments, our model outperforms the model of \cite{HermannK2014} in 107 experiments. This suggests that our model efficiently exploits the pivot language to facilitate cross language learning between other languages.  

Finally, we present the results for a monolingual classification task in Table \ref{table:mono}. The idea here is to see if learning common representations for multiple views can also help in improving the performance of a task involving only one view. \newcite{HermannK2014} argue that a Naive Bayes (NB) classifier trained using a bag-of-words representation of the documents is a very strong baseline. In fact, a classifier trained on document representations learned using their model does not beat a NB classifier for the task of monolingual classification. Rows 2 to 5 in Table \ref{table:mono} show the different settings tried by them (we refer the reader to \cite{HermannK2014}  for a detailed description of these settings). On the other hand our model is able to beat NB for 5/11 languages. Further, for 4 other languages (German, French, Romanian, Russian) its performance is only marginally poor than that of NB. 



\section{Experiment 2: Cross modal access using a pivot language}
In this experiment, we are interested in retrieving images given their captions in French (or German) and vice versa. However, for training we do not have any parallel data containing images and their French (or German) captions. Instead, we have the following datasets: (i) a dataset $\mathcal{Z}_{_{1}}$ containing images and their English captions and (ii) a dataset $\mathcal{Z}_{_{2}}$ containing English and their parallel French (or German) documents. For $\mathcal{Z}_{_{1}}$, we use the training split of MSCOCO dataset which contains 118K images and their English captions (see Section  \ref{sec:mscoco_data}). For $\mathcal{Z}_{_{2}}$, we use the English-French (or German) parallel documents from the train split of the TED corpus  (see Section \ref{sec:ted_data}). We use English as the pivot language and train Bridge Corrnet using $\mathcal{Z} = \{\mathcal{Z}_{_{1}}, \mathcal{Z}_{_{2}}\}$ to learn common representations for images, English text and French (or German) text. For text, we use bag-of-words representation and for image, we use the 4096 (fc6) representation got from a pretrained ConvNet (BVLC Reference CaffeNet \cite{jia2014caffe}). We learn hidden representations of size $D=200$ by training Bridge Corrnet for 20 epochs using stochastic gradient descent with mini-batches of size $20$. Each mini-batch contains data from only one of the  $\mathcal{Z}_{_{i}}s$.

For the task of retrieving captions given an image, we consider the 1000 images in our test set (see section \ref{sec:mscoco_data}) as queries. The 5000 French (or German) captions corresponding to these images (5 per image) are considered as documents. The task is then to retrieve the relevant captions for each image. We represent all the captions and images in the common space as computed using Bridge Corrnet. For a given query, we rank all the captions based on the Euclidean distance between the representation of the image and the caption. For the task of retrieving images given a caption, we simply reverse the role of the captions and images. In other words, each of the 5000 captions is treated as a query and the 1000 images are treated as documents. $\lambda$ was tuned to each task using a training/validation split. For the task of retrieving French/German captions given an image, $\lambda$ was tuned using the performance on the validation set for retrieving French (or German) sentences for a given English sentence. For the other task,  $\lambda$ was tuned using the performance on the validation set for retrieving images, given English captions. We do not use any image-French/German parallel data for tuning the hyperparameters.

We use recall@k as the performance metric and compare the following methods in Table 4:

\noindent \textbf{1. En-Image CorrNet}: This is the CorrNet model trained using only $\mathcal{Z}_{_{1}}$ as defined earlier in this section. The task is to retrieve English captions for a given image (or vice versa). This gives us an idea about the performance we could expect if direct parallel data is available between images and their captions in some language. We used the publicly available implementation of CorrNet provided by \cite{corrnet}.

\noindent \textbf{2. Bridge CorrNet}: This is the Bridge CorrNet model trained using $\mathcal{Z}_{_{1}}$ and $\mathcal{Z}_{_{2}}$ as defined earlier in this section. The task is to retrieve French (or German) captions for a given image (or vice versa).

\begin{table*}[htbp]
\centering
\begin{scriptsize}
\begin{tabular}{l | l | ccc | ccc}
\hline
&&\multicolumn{3}{c|}{\textbf{I To C}}& \multicolumn{3}{c}{\textbf{C To I}} \\
\textbf{Model}&\textbf{Captions}& \textbf{Recall@5} & \textbf{Recall@10} & \textbf{Recall@50} &  \textbf{Recall@5} & \textbf{Recall@10} & \textbf{Recall@50} \\\hline
En-Image CorrNet&English&0.118&0.190&0.456&0.091&0.168&0.532\\\hline
Bridge MAE&French&0.008&0.017&0.069 &0.007&0.013&0.063\\ 
2-CorrNet&French&0.018&0.024&0.085& 0.027&0.055&0.205\\ 
Bridge CorrNet&French&0.072&0.135&0.335& 0.032&0.060&0.235\\
CorrNet+MT&French&0.101&0.163&0.414& 0.069&0.127&0.416\\\hline

Bridge MAE&German&0.005&0.009&0.053& 0.006&0.013&0.058\\ 
2-CorrNet&German&0.009&0.013&0.071& 0.012&0.023&0.098\\ 
Bridge CorrNet&German&0.063&0.105&0.298& 0.027&0.049&0.183\\
CorrNet+MT&German&0.084&0.163&0.420& 0.061&0.107&0.343\\\hline
Random& & 0.006&0.009&0.044& 0.005&0.009&0.050\\\hline
\end{tabular}
 \caption{Performance of different models for image to caption (I to C) and caption to image (C to I) retrieval}
\end{scriptsize}
 \label{table:image_results}
\end{table*}

\noindent \textbf{3. Bridge MAE}: The Multimodal Autoencoder (MAE) proposed by \cite{ngiam11} was the only competing model which was easily extendable to the bridge case. We train their model using $\mathcal{Z}_{_{1}}$ and $\mathcal{Z}_{_{2}}$ to minimize a suitably modified objective function. We then use the representations learned to retrieve French (or German) captions for a given image (or vice versa).

\noindent \textbf{4. 2-CorrNet}: Here, we train two individual CorrNets using $\mathcal{Z}_{_{1}}$ and $\mathcal{Z}_{_{2}}$ respectively. For the task of retrieving images given a French (or German) caption we first find its nearest English caption using the Fr-En (or De-En) CorrNet. We then use this English caption to retrieve images using the En-Image CorrNet. Similarly, for retrieving captions given an image we use the En-Image CorrNet followed by the En-Fr (or En-De) CorrNet. 

\noindent \textbf{5. CorrNet + MT}: Here, we train an En-Image CorrNet using $\mathcal{Z}_{_{1}}$ and an Fr/De-En MT system\footnote{\url{http://www.statmt.org/moses/}} using $\mathcal{Z}_{_{2}}$. For the task of retrieving images given a French (or German) caption we translate the caption to English using the MT system. We then use this English caption to retrieve images using the En-Image CorrNet. For retrieving captions given images, we first translate all the 5000 French (or Germam) captions to English. We then embed these English translations (documents) and images (queries) in the common space computed using Image-En CorrNet and do a retrieval as explained earlier.

\noindent \textbf{6. Random}: A random image is returned for the given caption (and vice versa).

\begin{table*}[!t]
\centering

\begin{scriptsize}
\begin{tabular}{l@{\hskip 0.05in}l}
\multirow{5}{*}{ \includegraphics[height=1.9cm,width=2cm]{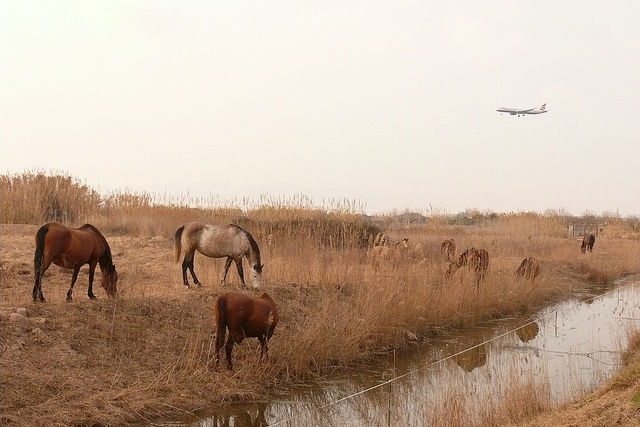}  } &1. Zwei Pferde stehen auf einem sandigen Strand nahe dem Ocean. (Two horses standing on a sandy beach near the ocean.)    \\
 &2. grasende Pferde auf einer trockenen Weide bei einem Flughafen. (Horses grazing in a dry pasture by an airport.)    \\
 & {\begin{tabular}[c]{@{}l@{}} \foreignlanguage{german}{3. ein Elefant , Wasser aufseinen Rückend sprühend , in einem staubigen Bereich neben einem Baum.}\\ (A elephant spraying water on its back in a dirt area next to tree .)\end{tabular}}\\
 
 & \foreignlanguage{german}{4. ein braunes pferd ißt hohes gras neben einem behälter mit wasser.}   (Brown horses eating tall grass beside a body of water .)  \\
 &5. vier Pferde grasen auf ein Feld mit braunem gras. (Four horses are grazing through a field of brown grass.)    \\

 &  \\
 
 \multirow{5}{*}{ \includegraphics[height=1.9cm,width=2cm]{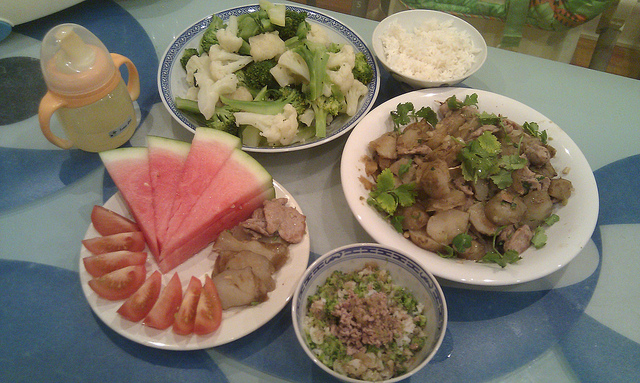} }
 &{\begin{tabular}[c]{@{}l@{}}1. Ein Teller mit Essen wie Sandwich , Chips , Suppe und einer Gurke.\\ (Plate of food including a sandwich , chips , soup and a pickle.)\end{tabular}}  \\
 &{\begin{tabular}[c]{@{}l@{}}\foreignlanguage{german}{2. Teller , gefüllt mit sortierten Früchten und Gemüse und einigem Fleisch.}\\(Plates filled with assorted fruits and veggies and some meat.)\end{tabular}}\\
 & \foreignlanguage{german}{3. Ein Tisch mit einer Schüssel Salat und einem Teller Pizza.} (a Table with a bowl of salad and plate with a cooked pizza .)    \\
 &4. Ein Teller mit Essen besteht aus Brokkoli und Rindfleisch. (A plate of food consists of broccoli and beef.)    \\
 & \foreignlanguage{german}{5. Eine Platte mit Fleisch und grünem Gemüse gemixt mit Sauce.}  (A plate with meat and green veggies mixed with sauce.)    \\
 
 &  \\
  
 \multirow{5}{*}{ \includegraphics[height=1.9cm,width=2cm]{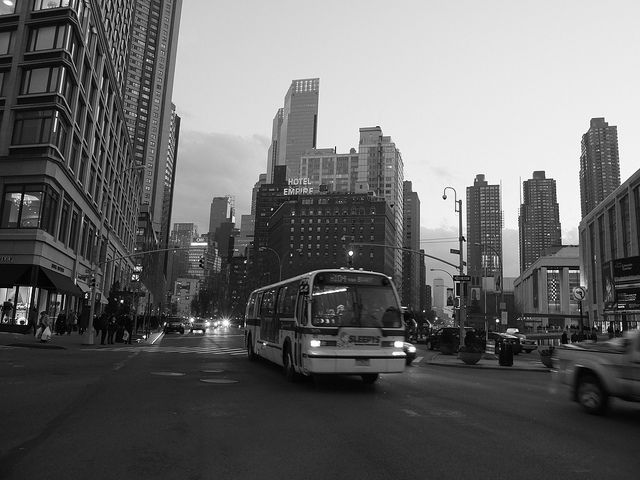}} 
 & {\begin{tabular}[c]{@{}l@{}}\foreignlanguage{french}{1. un bus de la conduite en ville dans une rue entourée par de grands immeubles.}\\  (A city bus driving down a street surrounded by tall buildings.)\end{tabular}}\\

 & {\begin{tabular}[c]{@{}l@{}}\foreignlanguage{french}{2. un bus de conduire dans une rue dans une ville avec des bâtiments de grande hauteur.}\\ (A bus driving down a street in a city with very tall buildings.)\end{tabular}}\\

 & \foreignlanguage{french}{3. bus de conduire dans une rue de ville surpeuplée.} (Double - decker bus driving down a crowded city street.) \\
 & \foreignlanguage{french}{4. le bus conduit à travers la ville sur une rue animée.} (The bus drives through the city on a busy street.)\\
 & \foreignlanguage{french}{5. un grand bus  coloré  est arrêté dans une rue de la ville.} (A big , colorful bus is stopped on a city street.)\\
 
  &  \\
  
 \multirow{5}{*}{ \includegraphics[height=1.9cm,width=2cm]{images/999.jpg}} 
 & {\begin{tabular}[c]{@{}l@{}}\foreignlanguage{french}{1. Un homme portant une batte de baseball à deux mains lors d'un jeu de balle professionnel.}\\  (A man holding a baseball bat with two hands at a professional ball game.)\end{tabular}}\\

 & \foreignlanguage{french}{2. un joueur de tennis balance une raquette à une balle.} (A tennis player swinging a racket t a ball.) \\
 & \foreignlanguage{french}{3. un garçon qui est de frapper une balle avec une batte de baseball.} (A boy that is hitting a ball with a baseball bat.)\\
 & \foreignlanguage{french}{4. une équipe de joueurs de baseball jouant un jeu de base-ball.} (A team of baseball players playing a game of baseball.)\\
 & \foreignlanguage{french}{5. un garçon se prépare à frapper une balle de tennis avec une raquette.}  (A boy prepares to hit a tennis ball with a racquet.)\\

\end{tabular}
\end{scriptsize}
 \caption{Images and their top-5 nearest captions based on representations learned using Bridge CorrNet. First two examples show German captions and the last two examples show French captions. English translations are given in parenthesis.}
\label{table:caption_retrieval_examples}
\end{table*}

\begin{table*}[!t]
\hspace*{-0.6in}

\begin{scriptsize}
\begin{tabular}{l@{\hskip 0.1in}l@{\hskip 0.1in}l@{\hskip 0.1in}l@{\hskip 0.1in}l@{\hskip 0.1in}l}
{\begin{tabular}[b]{@{}l@{}}
\foreignlanguage{german}{Speisen und Getränke auf einem }\\ \foreignlanguage{german}{Tisch mit einer Frau essen im Hintergrund.}\\  (Food and beverages set on a table with\\ a woman eating in the background .) \end{tabular}} & {\includegraphics[height=1.8cm,width=1.9cm]{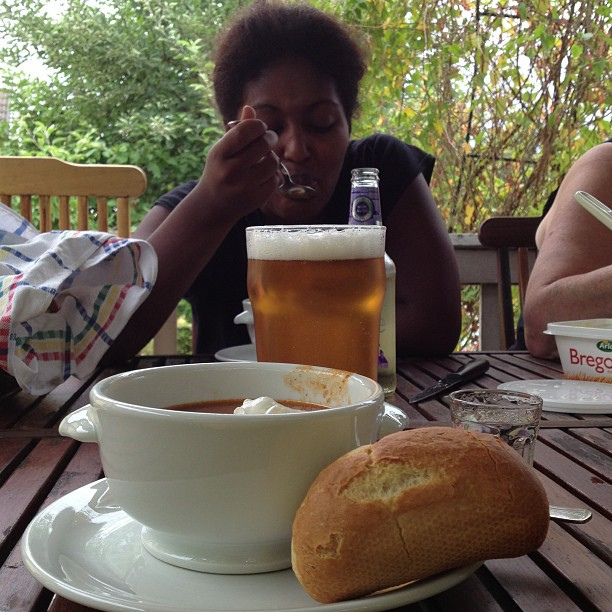}} & {\includegraphics[height=1.9cm,width=1.8cm]{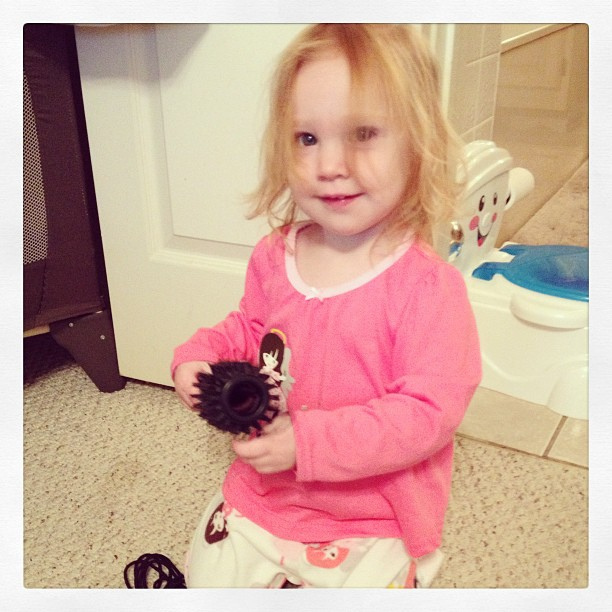}} & {\includegraphics[height=1.9cm,width=1.8cm]{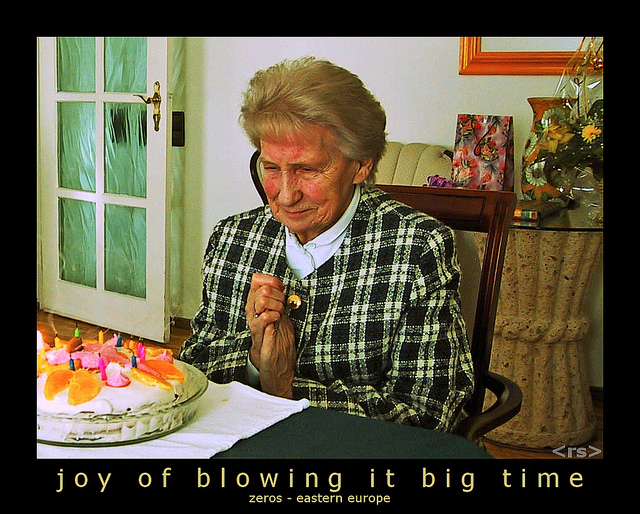}} & {\includegraphics[height=1.9cm,width=1.8cm]{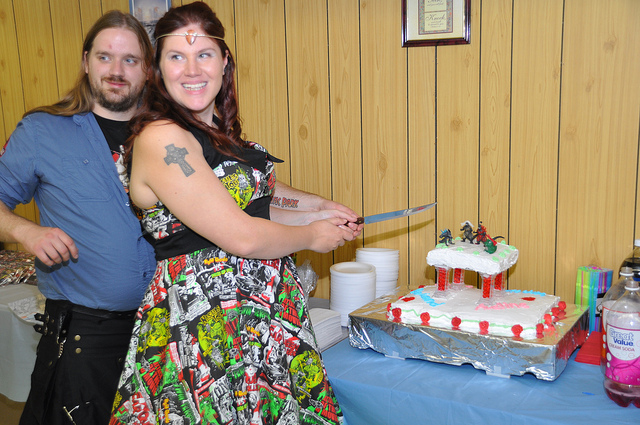}} &{\includegraphics[height=1.9cm,width=1.8cm]{images/4873e.jpg}} \\
{\begin{tabular}[b]{@{}l@{}}\foreignlanguage{german}{ein Foto von einem Laptop auf einem} \\ \foreignlanguage{german}{Bett mit einem Fernseher im Hintergrund.} \\ (A photo of a laptop on a bed with a tv \\in the background .)  \end{tabular}} & {\includegraphics[height=1.9cm,width=1.8cm]{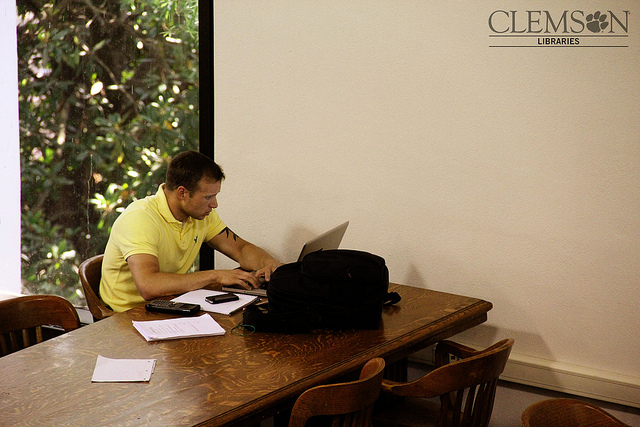}} & {\includegraphics[height=1.9cm,width=1.8cm]{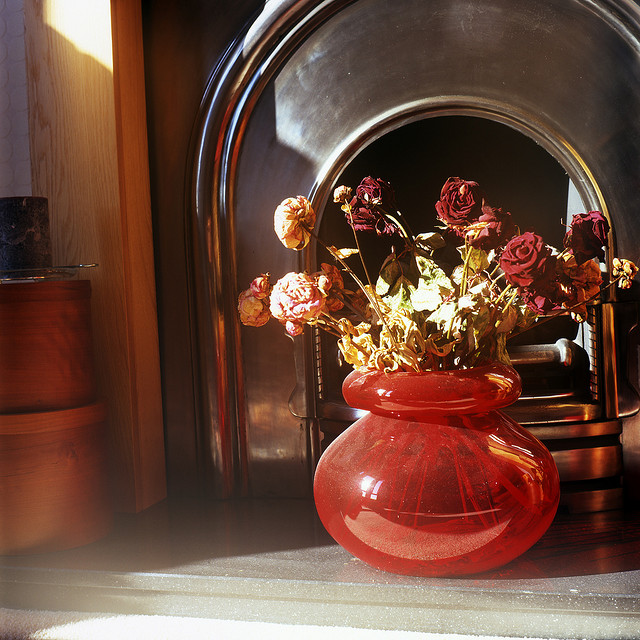}} & {\includegraphics[height=1.9cm,width=1.8cm]{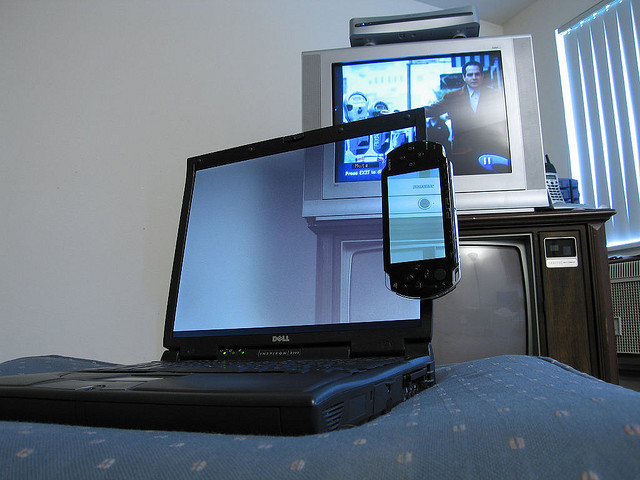}} & {\includegraphics[height=1.9cm,width=1.8cm]{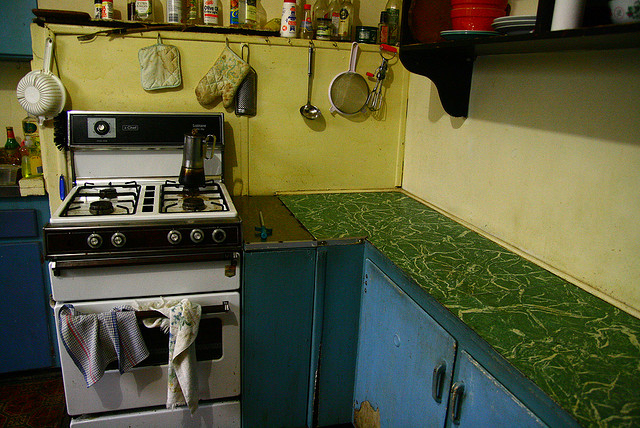}} &{\includegraphics[height=1.9cm,width=1.8cm]{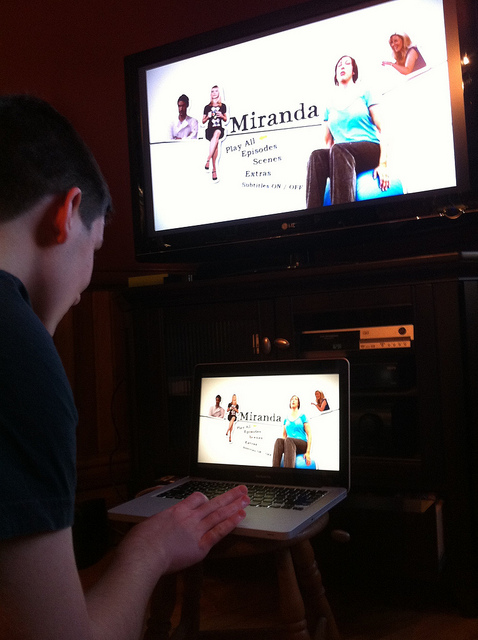}} \\
{\begin{tabular}[b]{@{}l@{}}\foreignlanguage{french}{un homme debout à côté de aa groupe de vaches.}\\ (A man standing next to a  group of cows.)
\end{tabular}} & {\includegraphics[height=1.8cm,width=1.9cm]{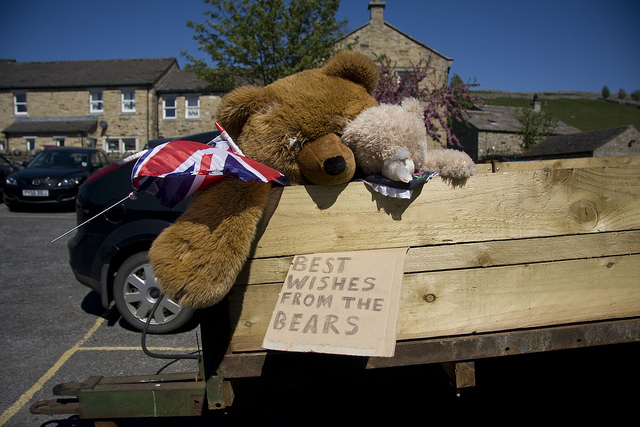}} & {\includegraphics[height=1.9cm,width=1.8cm]{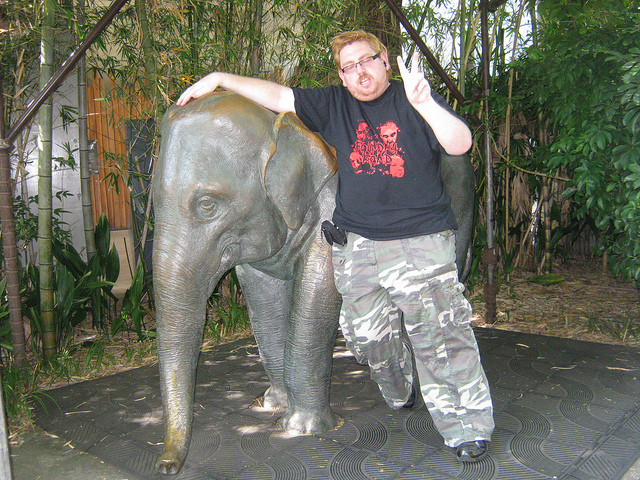}} & {\includegraphics[height=1.9cm,width=1.8cm]{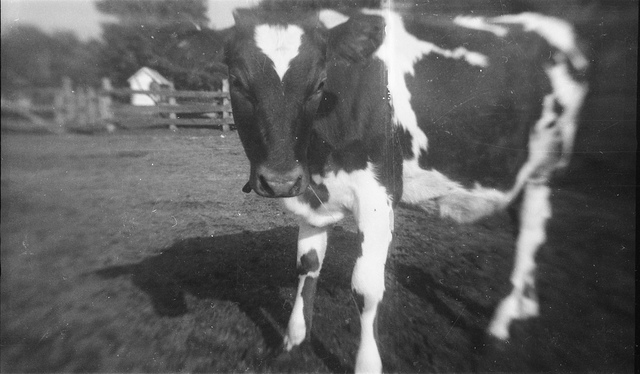}} & {\includegraphics[height=1.9cm,width=1.8cm]{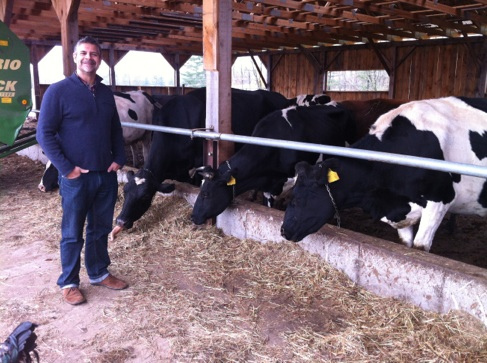}} &{\includegraphics[height=1.9cm,width=1.8cm]{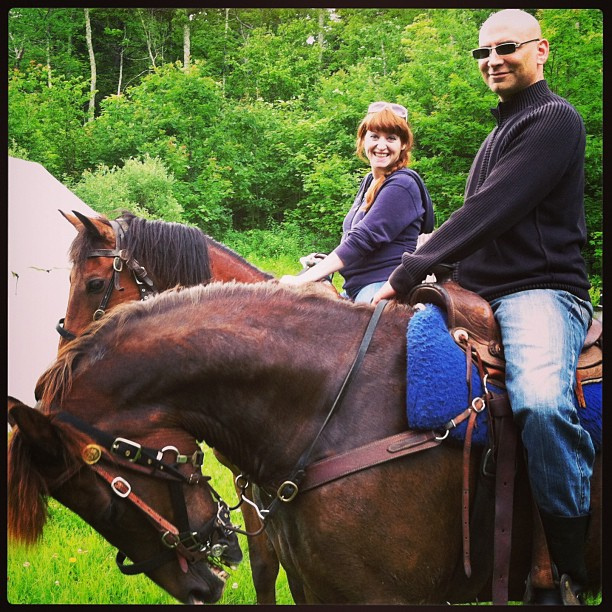}} \\
{\begin{tabular}[b]{@{}l@{}}\foreignlanguage{french}{personnes portant du matériel} \\
\foreignlanguage{french}{de ski en se tenant debout dans la neige.}
\\(People wearing ski equipment while \\standing in snow.) \end{tabular}} & {\includegraphics[height=1.9cm,width=1.8cm]{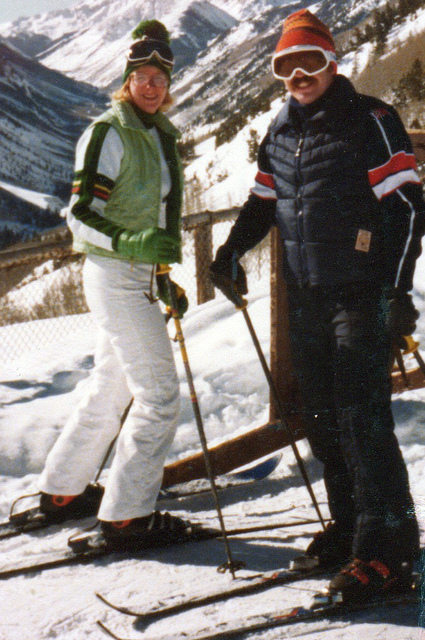}} & {\includegraphics[height=1.9cm,width=1.8cm]{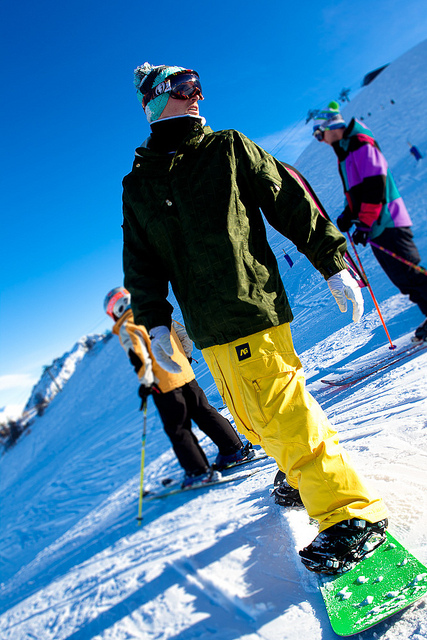}} & {\includegraphics[height=1.9cm,width=1.8cm]{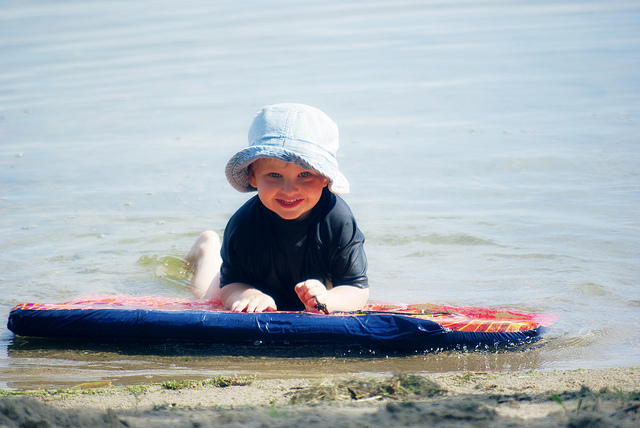}} & {\includegraphics[height=1.9cm,width=1.8cm]{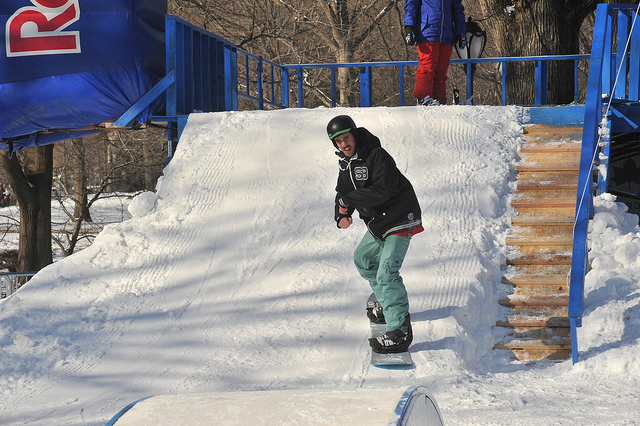}} &{\includegraphics[height=1.9cm,width=1.8cm]{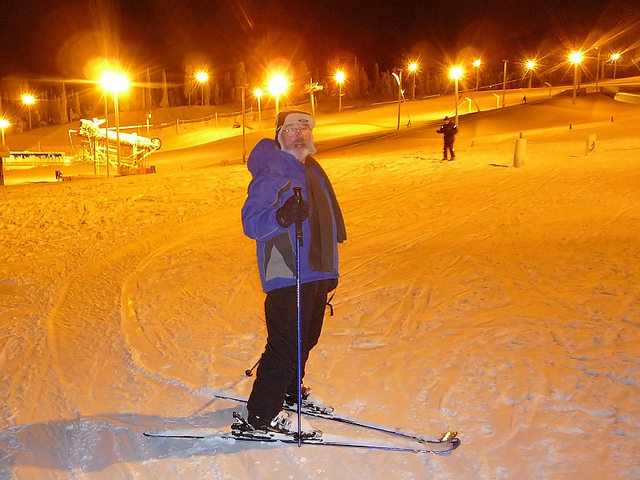}} \\

\end{tabular}
\end{scriptsize}
\caption{French and German queries and their top-5 nearest images based on representations learned using Bridge CorrNet. First two queries are in German and the last two queries are French. English translations are given in parenthesis.}
\label{table:image_retrieval_examples}

\end{table*}
From Table 4, we observe that \texttt{CorrNet + MT} is a very strong competitor and gives the best results. The main reason for this is that over the years MT has matured enough for language pairs such as Fr-En and De-En and it can generate almost perfect translations for short sentences (such as captions). In fact, the results for this method are almost comparable to what we could have hoped for if we had direct parallel data between Fr-Images and De-Images (as approximated by the first row in the table which reports cross-modal retrieval results between En-Images using direct parallel data between them for training). However, we would like to argue that learning a joint embedding for multiple views instead of having multiple pairwise systems is a more elegant solution and definitely merits further attention. Further, a ``translation system'' may not be available when we are dealing with modalities other than text (for example, there are no audio-to-video translation systems). In such cases, BridgeCorrNet could still be employed. In this context, the performance of BridgeCorrNet is definitely promising and shows that a model which jointly learns representations for multiple views can perform better than methods which learn pair-wise common representations (2-CorrNet). 




\subsection{Qualitative Analysis}
\label{sec:quali_analysis}
To get a qualitative feel for our model's performance, we refer the reader to Table \ref{table:caption_retrieval_examples} and \ref{table:image_retrieval_examples}. The first row in Table \ref{table:caption_retrieval_examples} shows an image and its top-5 nearest German captions (based on Euclidean distance between their common representations). As per our parallel image caption test set, only the second and fourth caption actually correspond to this image. However, we observe that the first and fifth caption are also semantically very related to the image. Both these captions talk about horses, grass or water body (ocean), \textit{etc}. Similarly the last row in Table \ref{table:caption_retrieval_examples} shows an image and its top-5 nearest French captions. None of these captions actually correspond to the image as per our parallel image caption test set. However, clearly the first, third and fourth caption are semantically very relevant to this image as all of them talk about baseball. Even the remaining two captions capture the concept of a \textit{sport} and \textit{raquet}. We can make a similar observation from Table \ref{table:image_retrieval_examples} where most of the top-5 retrieved images do not correspond to the French/German caption but they are semantically very similar. It is indeed impressive that the model is able to capture such cross modal semantics between images and French/German even without any direct parallel data between them.




\section{Conclusion}
In this paper, we propose Bridge Correlational Neural Networks which can learn common representations for multiple views even when parallel data is available only between these views and a pivot view. 
Our method performs better than the existing state of the art approaches on the cross language classification task and gives very promising results on the cross modal access task. We also release a new multilingual image caption benchmark (MIC benchmark) which will help in further research in this field\footnote{Details about the MIC benchmark and performance of various state-of-the-art models will be maintained at \url{http://sarathchandar.in/bridge-corrnet}}. 

\section*{Acknowledgments}

We thank the reviewers for their useful feedback. We also thank the workers from CrowdFlower for helping us in creating the MIC benchmark. Finally, we thank Amrita Saha (IBM Research India) for helping us in running some of the experiments. 

\bibliography{naaclhlt2016}
\bibliographystyle{naaclhlt2016}


\end{document}